\documentclass[sigconf]{aamas} 

\usepackage{balance} 
\usepackage{algorithm}
\usepackage{algorithmic}
\usepackage{caption}
\usepackage{subcaption}

\setcopyright{ifaamas}
\acmConference[AAMAS '21]{Proc.\@ of the 20th International Conference on Autonomous Agents and Multiagent Systems (AAMAS 2021)}{May 3--7, 2021}{Online}{U.~Endriss, A.~Now\'{e}, F.~Dignum, A.~Lomuscio (eds.)}
\copyrightyear{2021}
\acmYear{2021}
\acmDOI{}
\acmPrice{}
\acmISBN{}

\acmSubmissionID{679}

\title[AAMAS-2021 Formatting Instructions]{SEERL : Sample Efficient Ensemble Reinforcement Learning}

\author{Rohan Saphal}
\affiliation{
  \institution{Indian Institute of Technology Madras}
  \city{Chennai}
  \country{India}}
\email{rohansaphal@gmail.com}

\author{Balaraman Ravindran}
\affiliation{
  \institution{Robert Bosch Center for Data Science and Artificial Intelligence}
   \institution{Indian Institute of Technology Madras}
\city{Chennai}
  \country{India}}
\email{ravi@cse.iitm.ac.in }

\author{Dheevatsa Mudigere}
\affiliation{
  \institution{Facebook Inc}
  \city{Menlo Park}
  \country{USA}}
\email{dheevatsa@fb.com }

\author{Sasikant Avancha}
\affiliation{
  \institution{Intel Corporation}
    \city{Bangalore}
  \country{India}}
\email{sasikanth.avancha@intel.com }

\author{Bharat Kaul}
\affiliation{
  \institution{Intel Corporation}
    \city{Bangalore}
  \country{India}}
\email{bharat.kaul@intel.com }

\begin{abstract}
Ensemble learning is a very prevalent method employed in machine learning. The relative success of ensemble methods is attributed to their ability to tackle a wide range of instances and complex problems that require different low-level approaches. However, ensemble methods are relatively less popular in reinforcement learning owing to the high sample complexity and computational expense involved in obtaining a diverse ensemble. We present a novel training and model selection framework for model-free reinforcement algorithms that use ensembles of policies obtained from a single training run. These policies are diverse in nature and are learned through directed perturbation of the model parameters at regular intervals. We show that learning and selecting an adequately diverse set of policies is required for a good ensemble while extreme diversity can prove detrimental to overall performance. Selection of an adequately diverse set of policies is done through our novel policy selection framework. We evaluate our approach on challenging discrete and continuous control tasks and also discuss various ensembling strategies. Our framework is substantially sample efficient, computationally inexpensive and is seen to outperform state-of-the-art (SOTA) scores in Atari 2600 and Mujoco.
\end{abstract}


\newcommand{\BibTeX}{\rm B\kern-.05em{\sc i\kern-.025em b}\kern-.08em\TeX}

\begin{document}


\pagestyle{fancy}
\fancyhead{}


\maketitle 


\section{Introduction}
Deep reinforcement learning over the years has made considerable advancements with applications across a variety of domains -- from learning to play Atari 2600 suite from raw visual inputs \cite{mnih2015human}, mastering board games \cite{silver2017mastering,schrittwieser2019mastering}, learning locomotion skills for robotics \cite{schulman2015high,schulman2015trust,lillicrap2017continuous}, mastering Starcraft \cite{vinyals2019grandmaster}, the development of Alpha Fold \cite{evans2018novo} to predict the 3D structure of a protein and most recently, the improvements in model-based \cite{kaiser2019model} and off-policy reinforcement learning \cite{hessel2018rainbow,haarnoja2018soft}
\\
Nonetheless, it is a challenging task to create a single agent that performs well, is sample efficient and is robust. There are considerable hurdles surrounding training and optimization of deep reinforcement learning algorithms such as the sensitivity  to hyper-parameters, the high variance associated with the learning process and high sample complexity. In order to overcome these problems, we exploit the well-known concept of ensemble learning \cite{dietterich2000ensemble} and adapt it for reinforcement learning in a novel way. Traditionally, the idea of using ensembles in reinforcement learning settings is associated with combining multiple value functions or policies from different agents. These agents could be the same algorithm trained across different hyper-parameter settings, generated by different algorithms altogether \cite{wiering2008ensemble,duell2013ensembles,fausser2015neural} or by training multiple networks of an agent in parallel \cite{lee2020sunrise}. Training multiple such agents is an approach that cannot be used in practice owing to high sample complexity and computational expense. 


We tackle this problem by creating sufficiently diverse policies from a single training run. The policies are generated in a serial manner, one after the other, where the subsequent policies take into account the diversity from earlier policies. Our framework is seen to achieve state-of-the-art (SOTA) performance on popular reinforcement learning domains.  
\\
 Our approach to sequentially generate  policies is inspired by the recent developments  in the deep learning literature studying the effects of learning rate schedules and their impact on generalization \cite{li2019towards,nakkiran2020learning}. It is shown that learning rate annealing generalizes better than using a small constant learning rate and high initial learning rate impacts the model's selection of the local minimum to converge \cite{li2019towards, jastrzebski2020break}. We leverage these properties of neural network training to learn a diverse ensemble of policies.
The diversity among the policies are obtained by the directed perturbation of the model weights at regular intervals. The directed perturbation is induced by sudden and sharp variations in the learning rate, and for doing so, we employ cyclical learning rates \cite{loshchilov2016sgdr}. When the model weights are perturbed using larger learning rates,  the directed motion along the gradient direction prevents the optimizer from settling in any sharp basins and directs the model into the general vicinity of a local minima \cite{li2019towards}. Annealing the learning rates during training leads the optimizer to converge to some local minima and improves generalization \cite{nakkiran2020learning}. We leverage the diversity of the policies learned at these different local minima for the ensemble. We also show through experimentation that directed perturbation and not random perturbation is necessary for obtaining diverse policies. We also empirically show that an extremely diverse set of policies do not form a good ensemble.
\\
In order to prevent bias from sub-optimal policies in the ensemble, we introduce a novel framework that selects the best subset of policies to include in the ensemble. Our approach uses the trajectories obtained during training the policies to find this subset. 
\\
Since we use models from a single training run instead of training $M$ different models independently from scratch, we refer to our approach as Sample Efficient Ensemble Reinforcement Learning (SEERL).
 To summarize, our main contributions are:
\begin{itemize}
    \item A sample efficient framework for learning $M$ diverse models in a serial manner from a single training run with no additional computation cost. The framework can adopt any existing reinforcement learning algorithm to create an ensemble
    \item A novel optimization framework to select the best subset of policies for the ensemble. It acts as a filter to choose the best performing and diverse subset of policies without using any additional samples from the environment.
    \item Evaluation of various ensemble strategies for discrete and continuous action spaces with SOTA performance across multiple environments
\end{itemize}
We demonstrate the effectiveness of SEERL for discrete (Atari 2600 \cite{bellemare2013arcade}) and continuous control benchmarks (Mujoco \cite{todorov2012mujoco}). Our experiments show that SEERL consistently outperforms popular model-free and model-based  reinforcement learning methods including those that use ensembles.

\section{Related work}
There has recently been a small body of work on using ensembles for reinforcement learning, primarily the use of ensembles during the training phase to reduce the variance and improve robustness of the policy.
\\
Value function based methods such as Averaged DQN \cite{anschel2017averaged} train multiple Q networks in parallel with different weight initialization and average the Q values from all the different networks to reduce variance. It results in learning policies that are much more stable. However, the approach requires training multiple networks simultaneously and a possibility that the model might diverge if either of the Q values being averaged is biased.
\\
Bootsrapped DQN \cite{osband2016deep} uses an ensemble of Q functions for efficient exploration but does not enforce diversity in the Q networks. The Q head is chosen at random and used to generate actions for that episode. The multiple heads are also trained in parallel while we train our models in a serial manner. 
\\
SUNRISE \cite{lee2020sunrise} also uses an ensemble of Q networks that are randomly initialized and each head trained with a different set of samples to stabilize learning and enforce diversity. The actions are selected by examining the Upper Confidence Bound(UCB) of the Q values and choosing the action with the highest upper confidence bound to allow for exploration. However, the framework does not account for bias from the different Q heads which in turn would affect the UCB. Our policy selection framework prevents such biases by filtering out such models during evaluation. Additionally, SUNRISE works only where the base learner is an off policy algorithm whereas our framework works with both on-policy and off-policy methods.  
\\
Earlier works \cite{wiering2008ensemble,duell2013ensembles,fausser2015neural,fausser2015selective} explore the idea of value function ensembles and policy ensembles during evaluation phase. However, value function ensembles from different algorithms trained independently could degrade performance as they tend to converge to different estimates of the value function that are mutually inconsistent. An alternative method \cite{marivate2013ensemble} tries to tackle this problem by having a meta-learner linearly combine the value functions from different algorithms during training time to adjust for the inherent bias and variance. Although training multiple policies or value functions in parallel is sample efficient, it tends to reduce the diversity among them and is computationally expensive.
Our method combines the best of both approaches and improves the performance of the algorithm by balancing sample complexity with the computational expense.
Training deep neural network architectures with cycling learning rates \cite{smith2015no,loshchilov2016sgdr} and using ensembles in supervised learning settings, \cite{huang2017snapshot} have shown to be useful. The authors \cite{huang2017snapshot} show that in each cycle, the models obtained are comparable with those learned using traditional learning rate schedules. Even though the model is seen to degrade in performance temporarily, the new model surpasses the previous one, as the learning rate anneals.

\section{Preliminaries}
Reinforcement learning is associated with sequential decision making and involves the interaction of an agent with an environment. In this paper, we consider a discrete-time finite-horizon Markov Decision Process(MDP) defined by (${S}$, ${A}$, ${P}$, ${r}$, $\gamma$, $\rho$), where ${S}$ denotes the state space, ${A}$ denotes the action space, ${P}\equiv P(s'|s,a) $ the transition function, $\rho$, the probability distribution over the initial states, $r(s,a)=\mathbb{E}[R_{t+1}|s_t,a_t]$, the reward function with $R_{t+1}$ being the scalar reward at time $t+1$ when the agent in state $s_t$ took the action $a_t$,  and $\gamma \in (0,1)$ the discount factor. The policy dictates the behavior of an agent at a particular state in an environment. More formally, a policy is defined by $\pi :{S} \rightarrow \mathcal{P}({A})$ where $\mathcal{P}({A})$ denotes the probability distribution over actions $a$ $\epsilon$ ${A}$. The objective of the agent is to maximize the discounted return $G_t = \Sigma_{i=t}^{T}$  $\gamma^{i-t}$ $R_i$

\section{SEERL}
In this paper, we propose SEERL, a novel framework for learning and selecting an ensemble of diverse policies obtained from a single training instance. Unlike supervised learning, where the dataset can be reused for training different models, training multiple RL agents on the same trajectories can result in a lack of diversity among the policies. Instead, multiple RL agents can be trained independently for an ensemble but would suffer from high sample complexity. 
If each agent requires $N$ number of samples and the computational expense for training a single agent is $C$, then training $M$ agents independently require $M \times N$ samples and $M \times C$ computational cost. If trained in parallel, only $N$ samples are required, but the computational cost remains at $M \times  C$. 
Though training agents in parallel is a possible solution to tackle sample complexity, it is computationally expensive and limits the diversity among the learned policies since every policy observes the same state.
\\
SEERL follows a two-step process-learning policies and policy selection, summarized in Algorithm \ref{alg:algorithm}.
Learning policies involves saving $M$ policies during training at periodic intervals when the learning rate anneals to a small value.
This is followed by policy selection that finds the best subset of $m$ policies from the entire set $M$. Policy selection uses an optimization framework that uses the $N$ samples obtained during training to find the $m$ policies.
These policies are then used as an ensemble during the evaluation phase. 
Thus SEERL produces $m$ models for the ensemble, requiring only $N$ number of samples and a computational expense of $C$. In the later sections, we empirically verify that the policies from different local minima are diverse in nature.

\begin{figure}[t]
\centering
    \includegraphics[width=0.8\linewidth]{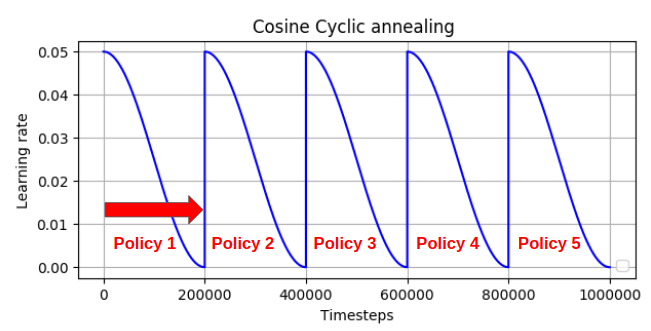}
    \caption{ Cyclical cosine annealing learning rate schedule. $\alpha_0$ is set at $0.05$, number of models $M=5$ and training timesteps $T=1000000$ }
    \label{clr}
    \Description{Cyclical cosine annealing learning rate schedule}
\end{figure}

\subsection{Learning policies}
 Learning rate annealing \cite{nakkiran2020learning, li2019towards} is shown to improve generalization and the use of high initial learning rate \cite{jastrzebski2020break} determines the local minima for the model. 
 We use cosine cyclical annealing learning rate schedule \cite{loshchilov2016sgdr} to introduce learning rate annealing and learn multiple policies, as shown in Figure \ref{clr}. Depending on the number of time-steps needed to train the agent, and the number of models needed for the ensemble, the learning rate schedule can be calculated. 
\\
As the learning rate anneals to a small value, the model converges to a local minimum, and we obtain the first policy. By increasing the learning rate, the model is perturbed along the gradient direction and dislodged from its local minima. In other words, if $M$ models are required, we split the training process into $M$ different training cycles where in each cycle the model starts at a high learning rate and anneals to a small value. The high learning rate is significant as it provides energy to the policy to escape the local minima and the learning rate annealing traps it into a well behaved local minima. The annealing helps in generalization and converging to a local minima while the directed perturbation re-orients the model to converge to a different local minima. The formulation is as follows:
\begin{eqnarray}
\label{eqn 1}
\alpha(t)=\frac{\alpha_0}{2}\left(\cos\left({\frac{\pi \mod{(t-1,\lceil T/M \rceil)}}{\lceil T/M \rceil}}\right)+ 1\right)
\end{eqnarray}%
where $\alpha_0$ is the initial learning rate, $t$ is the time-step, and $T$ is the total number of time steps for which the agent is trained, and $M$ is the number of models.

\subsection{Policy selection}
\label{section:4.2}
In order to avoid bias from the more inferior policies, the best $m$ policies should ideally be selected for the ensemble.
At the same time, the policies also need to be diverse to obtain good performance in different parts of the state space when used in an ensemble. Near identical policies would not yield much improvement in an ensemble. 

\subsubsection{\textbf{Framework} :}
We propose an optimization framework to select the best subset of policies. The formulation has two parts, a policy error term and a Kullback-Liebler (KL) divergence term indicative of diversity. Only optimizing for the policy error term would result in the selection of policies with excellent performance. The addition of the KL divergence term helps balance the requirements of performance and diversity.  We have a hyper-parameter $\beta \in [1,2)$, that balances between diversity and performance. The KL divergence is calculated based on the action distribution between the two polices over the state space. The formulation is as follows :
\begin{equation}
\label{eqn 2}
J(w)=\sum_{s \in S}{\Bar{P}(s)}\left[\sum_{i \in M} w_i B_{i}(s) \right]^2
\end{equation}
\begin{equation}
\label{eqn 3}
\resizebox{0.91\linewidth}{!}{$
    \displaystyle
B_{i}(s)=\left(\sum_{a \in A} L(s,a) -\frac{\beta}{M-1} \sum_{k \in M, k \neq i} \text{KL}(\pi_i (s) || \pi_k(s))\right)
$}
\end{equation}
with the following constraints $\sum_{i \in M}w_{i} =1, w_{i}\geq 0   \forall i $.
$S$ is the set of states and $\Bar{P}(s)$ is the probability of observing a particular state. $L(s, a)$ is the weighted error associated with the model and is indicative of the performance of the model. We weigh the loss in a manner that we give more weight when the loss is above a certain threshold value, and the action taken by the model, $a$, matches the action taken by the ensemble $a_e$. We formalize $L(s,a)$ as follows :
\begin{equation}
\resizebox{0.91\linewidth}{!}{$
\label{eqn 4}
    \displaystyle
  L(s,a)=\left\{
  \begin{array}{@{}ll@{}}
    1, & \text{if} \ |L'(s,a)| \geq T_{err}\ \textbf{and}  \ |a-a_e|<\epsilon,\ \text{a is continuous}  \\
        1, & \text{if} \ |L'(s,a)| \geq T_{err}\ \textbf{and}  \ a=a_e, \ \text{a is discrete} \\
    0, & \text{otherwise}
  \end{array}\right.
  $}
\end{equation} 
$L'(s, a)$ is the total error for a particular state-action pair. While the overall process is algorithm agnostic, the details of the total error depend on the actual RL algorithm adopted. $T_{err}$ is the threshold value that is used to distinguish the right actions from the wrong. If $L'(s, a)$ is above the threshold error and the actions match, then we would like to consider that action from the policy as a wrong one.
\\
In a discrete action space, it is relatively easy to determine if the action taken by the model and ensemble are the same.
However, in continuous action space, the final ensemble action and the action taken by the policy might not coincide. Therefore, we introduce a  $\epsilon$ bound on the ensembled action. If the action from the model is within a $\epsilon$ distance from the ensembled action, we consider it as a match. $\epsilon$ ranges between 0.005 to 0.01, depending on the environment. 
\\ We use a squared loss formulation to capture the inter-dependencies among the policies. Instead, if the degree were 1, the objective function would be the weighted sum of the loss, and the one with the lowest error would be the best policy. By having a higher degree, we are capturing the dependencies among the policies. The dependencies arise because the initialization of a new policy is  at the termination of the previous policy and the policy network weights are shared.
\\
Let us consider the computation of the total error for the case of A2C, the total error, $L'(s, a)$,  for a state action pair is the weighted sum of the policy gradient loss and the value function loss. $V(s)$ and $A(s,a)$  is the value function and the advantage function respectively at state $s$ obtained by using policy $\pi(a|s)$ .
\begin{equation}
\label{eqn 5}
V_{loss}(s) = r(s,a)+\gamma V(s')-V(s)
\end{equation} 
\begin{equation}
\label{eqn 6}
\pi_{loss}(a|s)=-log(\pi(a|s)) A(s,a)
\end{equation} 
\begin{equation}
\label{eqn 7}
L'(s,a)=\pi_{loss}(a|s) + V_{loss}(s) * C_v
\end{equation} 
\\
$\gamma$ is the discount factor, $C_v$ is the coefficient used for weighting the value function against the policy gradient loss. 
\\
By minimizing this objective function, we obtain the values of $w_i$, the Lagrange multipliers to this optimization framework. To choose the best ensemble, we select the $m$ policies with the highest corresponding lagrange multipliers.

\subsubsection{\textbf{Implementation detail} :}
In order to solve the optimization problem, we can re-frame it as a  quadratic programming problem with box constraints as follows:
\begin{equation}
\label{eqn 8}
J(w)=w^{T}Bw
\end{equation}
and,
\begin{equation}
\label{eqn 9}
B_{ij}=\sum_{s \in S}{\Bar{P} (s)}b_{i}b_{j}
\end{equation}
where,
\begin{equation}
\label{eqn 10}
\resizebox{0.91\linewidth}{!}{$
    \displaystyle
b_{i}=\left[\sum_{a \in A} L(s,a) -\frac{\beta}{M-1} \sum_{k \in M, k \neq i} \text{KL}(\pi_i (s) || \pi_k(s))\right]
$}
\end{equation}
The formulation is still subjected to all the linear constraints as earlier. The matrix $B$ is a positive definite matrix since it is an inner product of the loss terms. This results in a convex objective function for which the global minimum can be found.
In order to run this optimization, we select the states from multiple trajectories in the training samples. This framework efficiently reuses data and is better in comparison to evaluating the policies directly in the environment.
Without this framework, the method to find the best ensemble subset would be to evaluate all possible combinations.
For comparison, if each ensemble model is evaluated using $N'$ samples from the environment, a total of $m \times N'$ samples will be used up in selecting the $m$ policies.

\begin{algorithm}[tb]
\caption{SEERL}
\label{alg:algorithm}
\begin{algorithmic}[1]
\STATE {\bfseries Input} Initialize a policy $\pi_\theta$, training time-steps $T$, evaluation time-steps $T'$, number of policies $M$, maximum learning rate $\alpha_0$, number of policies to ensemble $m$, ensemble strategy $E$\\
\STATE {\bfseries Output} Average reward during evaluation\\
\STATE {\bfseries Training}
\WHILE{$t \leq T$}
\STATE Calculate the learning rate, based on the inputs to the cosine annealing learning rate schedule $f$
\STATE $\alpha(t)=f( \alpha_0, t, T, M)$ //Equation \ref{eqn 1}
\STATE Train the agent using $\alpha(t)$
\IF {t $\bmod$ (T/M)}
\STATE Save policy $\pi_\theta^i$ for $i={1,2\dots ,M}$
\ENDIF
\ENDWHILE
\STATE {\bfseries Evaluation}

\STATE Select the $m$ policies using the Policy selection process //Section \ref{section:4.2}
\STATE Select an ensemble strategy $E$
\WHILE{$t \leq T'$}
\STATE Collect actions from the $m$ policies, ${a_1,a_2\dots, a_m}$ for environment state $s_t$
\STATE Find the optimal action, $a^*$ using $E$ //Section \ref{section:4.3}
\STATE Perform action $a^*$ on the environment 
\STATE Obtain cumulative reward for the episode, $r_t$ and the next state $s_{t+1}$
\ENDWHILE
\STATE \textbf{return} Average reward obtained during evaluation
\end{algorithmic}
\end{algorithm}
\subsection{Ensemble techniques}
\label{section:4.3}
Once the $m$ policies are chosen, depending on the complexity of the action space, discrete or continuous, there are multiple strategies to ensemble the actions in the environment.
We divide the ensemble strategy into two categories, for discrete and continuous action spaces.
\subsubsection{\textbf{Ensemble in discrete action spaces :}}
In discrete action spaces, we consider majority voting as a good solution.
Due to different fixed point convergences of value functions of algorithms trained independently, it is not possible to compare actions by their $Q$ values.
\begin{equation}
\label{eqn 11}
\pi(a|s)=\mathop{argmax}_{a \in A(s)}\left[\sum_{i \in m} N_i(s,a)\right]
\end{equation}%
where $ N_i(s,a)$ is one if the agent $i$ takes action $a$ in state $s$, else zero.
It the case of a tie, random action is chosen among the set of actions having the tie.
\begin{figure*}[t]
        \centering
        \begin{subfigure}{0.48\textwidth}
            \includegraphics[width=\textwidth]{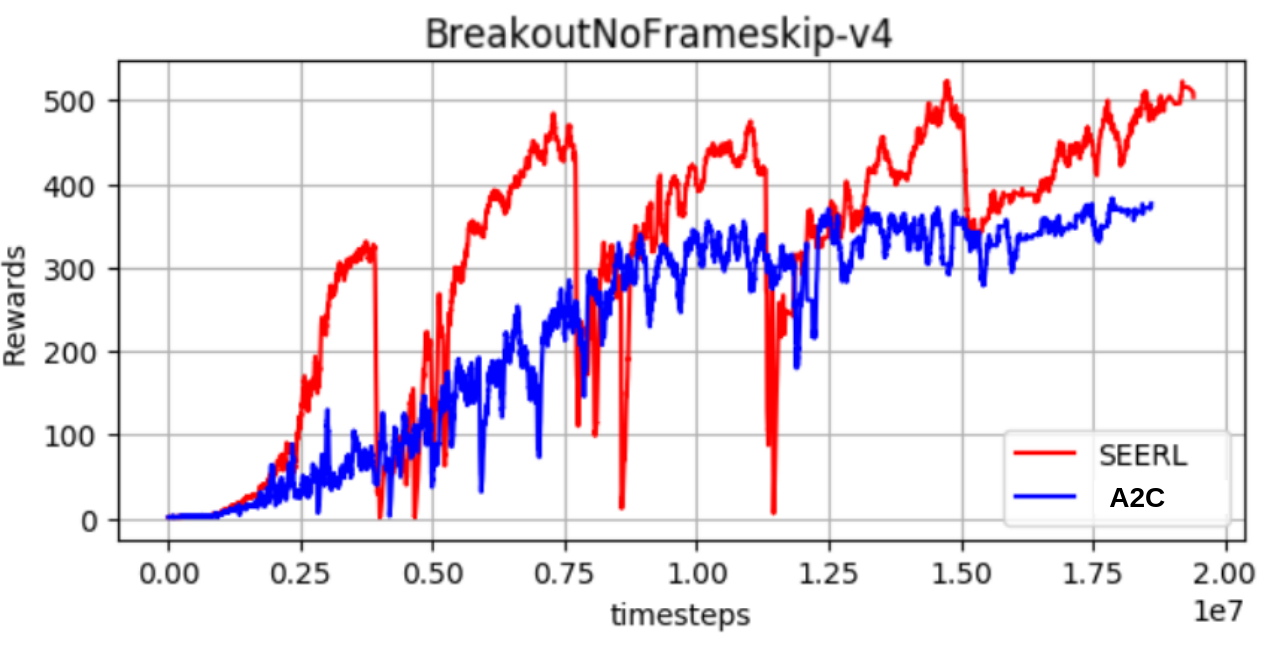}
            \caption{Training performance of SEERL vs A2C on Breakout }
            \label{fig:gull1}
        \end{subfigure}
        \hfill
        \begin{subfigure}{0.48\textwidth}
            \includegraphics[width=\textwidth]{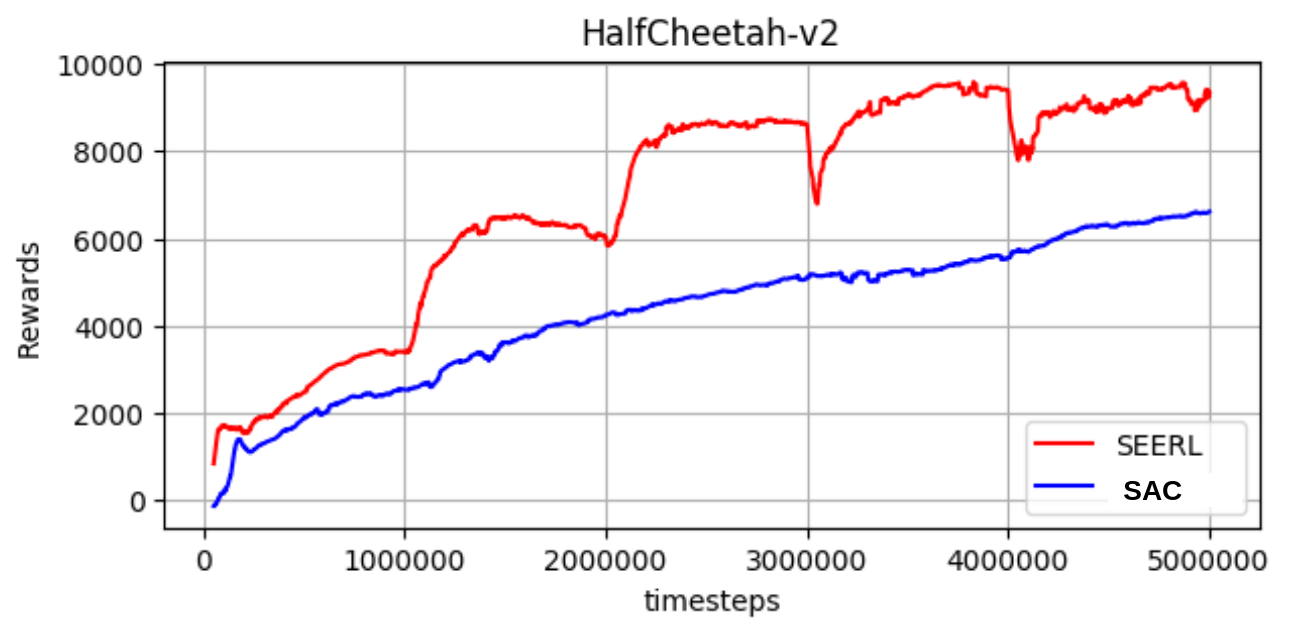}
            \caption{ Training performance of SEERL vs SAC on Half-Cheetah}
            \label{fig:gull2}
        \end{subfigure}
        \hfill
        \begin{subfigure}{0.3\textwidth}
            \includegraphics[width=\textwidth]{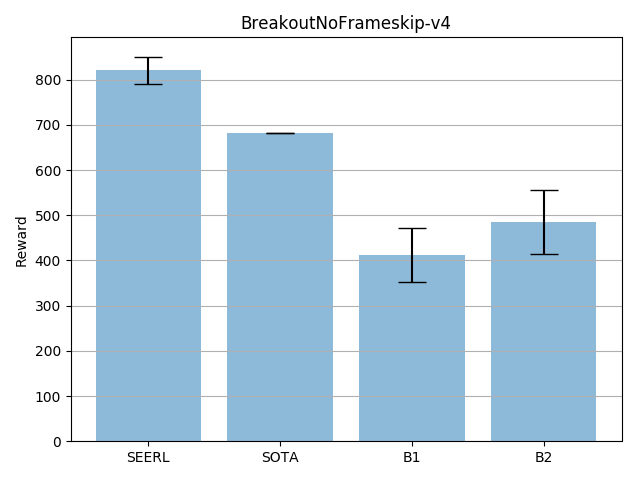}
            \caption{Comparison between SEERL(A2C), SOTA and baseline ensembles B1(A2C), B2(A2C and ACER) on Breakout }
            \label{fig:gull3}
        \end{subfigure}
        \hfill
        \begin{subfigure}{0.3\textwidth}
            \includegraphics[width=\textwidth]{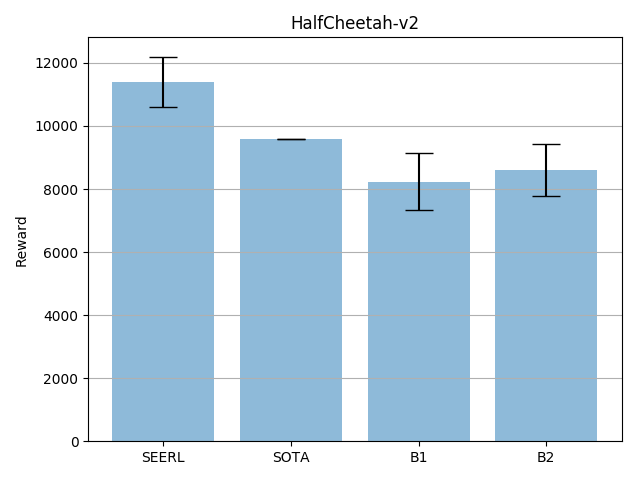}
            \caption{Comparison between SEERL(SAC), SOTA and baseline ensembles B1(SAC), B2(SAC and DDPG) for Half Cheetah }
            \label{fig:gull4}
        \end{subfigure}
        \hfill
        \begin{subfigure}{0.3\textwidth}
            \includegraphics[width=\textwidth]{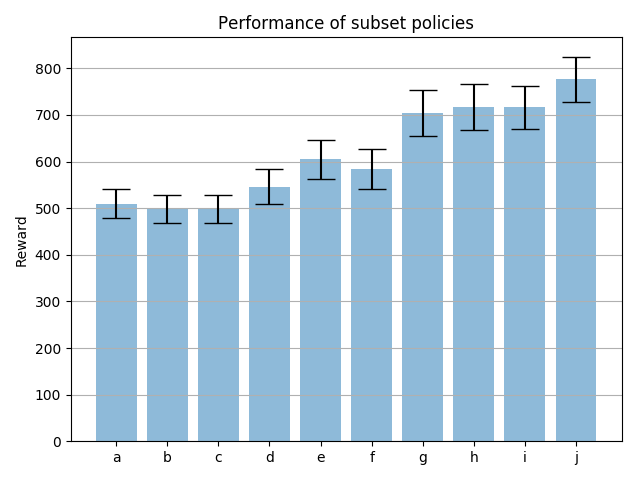}
            \caption{Performance of each subset of policies , an ensemble of 3 policies selected from a set of 5, on Breakout }
            \label{fig:gull5}
        \end{subfigure}
\end{figure*}

\subsubsection{\textbf{Ensemble in continuous action spaces :}}
In continuous action spaces, \cite{duell2013ensembles} proposes multiple strategies to find the optimal action. However, the performance comparison of the strategies is not provided and environments considered are too simple. The different strategies are as follows:
\begin{itemize}
    \item Averaging: We take the average of all the actions as part of the ensemble. This strategy could fail in settings where one or more of the actions are extremely biased and thereby shifts the calculated value away from the true mean value.
    \item Binning: This is the equivalent of majority voting in a continuous action space setting. We discretize the action space into multiple bins of equal size and average the bin with the most number of actions. The average value obtained is the optimal action to take. Through this method, we have discretized the action space, sorted the bins based on its bin-count, and calculated the mean of the bin with the highest bin-count. The hyper-parameter to be specified here is the number of bins. We use five bins in our experiments
    \item Density-based Selection(DBS): This approach tries to search for the action with the highest density in the action space. Given $M$ action vectors, $a$, each of $k$ dimensions to be ensembled, we calculate the density of each action vector using Parzen windows as follows:
    \begin{equation}
    \label{eqn 12}
    d_{i}=\sum_{j=1}^{M}exp{-\frac{\sum_{l=1}^{k}(a_{il}-a_{jl})^{2}}{h^{2}}}
    \end{equation}%
    
     The action with the highest density, $d_i$, is selected as the final action. The only parameter to be specified is $h$, the window width, and we have chosen $h=0.0001$ in our experiments. 
    \item Selection through Elimination(STE): This approach eliminates action based on the Euclidean distance. We calculate the mean of the action vectors and compute the euclidean distance to each action from the mean. The action with the largest euclidean distance is eliminated, and the mean is re-computed. The process is repeated until two actions remain. The final action is chosen as the average of the two actions.  
\end{itemize}

\begin{table*}[t]
\centering
\caption{Evaluation scores across Atari 2600 games averaged over 100 episodes. We report the published scores for DQN \cite{mnih2015human}, Distributional DQN\cite{bellemare2017distributional} and Rainbow DQN. SEERL uses A2C as base learner is the best performing of the methods, topping in 11 of the 15 games. All results represent the average of 5 random training runs. We run each game for 20M environment time steps.}
\begin{tabular}{rlllll}\toprule
{Game}&  {DQN} & A2C &Distrib. DQN  &  Rainbow DQN  & SEERL  \\\midrule
Alien &  634 & 518.4 & 1997.5 & \textbf{6022.4} & 1924\\
Assault &  178.4 & 263.9 & 237.7 & 202.8 & \textbf{341.2}\\
Bank Heist &  312.7 & 970.1 & 835.6 & 826 & \textbf{1124.1}\\
Battle zone &  23750 & 12950 & 32250 & \textbf{52040} & 28580\\
Breakout &  354.0 & 681.9 & 584.9 & 379.5 & \textbf{821}\\
Freeway &  26.9 & 0.1 & 28.8 & 29.1 & \textbf{33.1}\\
Frostbite&  496.1 & 190.5 & {2813.9} & \textbf{4141.1} & 1226\\
Krull &  6206.0& 5560.0 & 6757.8 & {6715.5} & \textbf{6795.2}\\
MsPacman &  1092.3 & 653.7 & 2064.1 & {2570.2} & \textbf{2614.2}\\
Pong &  18 & 5.6 & 18.9 & 19.1 & \textbf{19.8}\\
PrivateEye &  207.9 & 206.9 & \textbf{5717.4} & 1704.4 & {722.1}\\
Qbert &  9271.5 & 15148.8 & 15035.9 & 18397.6 & \textbf{18834.2}\\
Road Runner &  35215 & 34216 & {56086} & 54261 & \textbf{58624.2}\\
Robotank &  58.7 & 32.8& 49.8 & 55.2 & \textbf{61.2}\\
Seaquest &  4216.7 & 2355.4 & 3275.4 & \textbf{19176.0} & {4811.2}\\
\bottomrule
\end{tabular}
\label{tab1}
\end{table*}

\begin{table}[t]
\centering
\caption{Evaluation scores across Mujoco environments averaged over 100 episodes. SEERL using SAC as base learner is seen to outperform across 5 of the 6 environments. All results represent the average of 5 random training runs.}
\begin{tabular}{rlllll}\toprule
Environment &  TRPO  & PPO  &DDPG  &  SAC & SEERL  \\
\midrule
Ant &  2342.2 & 962 & 342.4 & 1958 & \textbf{2564}\\
Half Cheetah &  4233 & 1820 & 5440 &7269 & \textbf{11658}\\
Hopper & 2252 & 1112 & 1233 & \textbf{3379.2} & 3156\\
Humanoid &  3882 & 735 & 101.4 & 4380 & \textbf{4845}\\
Swimmer &  121.2& 42.4 & 43.4 & 44.6& \textbf{162.4}\\
Walker 2D&  3215 & 1892 & 782 & 2112 & \textbf{4366}\\
\bottomrule
\end{tabular}

\label{tab2}
\end{table}

\begin{table*}[t]
\centering
\caption{Evaluation scores across Atari 2600 games averaged over 10 episodes using Rainbow DQN \cite{hessel2018rainbow} as base learner for SEERL at 100K interactions. We report the published scores for SimPLe \cite{kaiser2019model}, CURL\cite{srinivas2020curl} and SUNRISE\cite{lee2020sunrise} at 100K interactions. All results represent the average of 3 random training runs.}
\begin{tabular}{rlllll}\toprule
Game &   Rainbow DQN  & SimPLe  &CURL  &  SUNRISE  & SEERL \\
\hline
Alien         &  789   & 616.9 & 558.2 & \textbf{872} & 800\\
Amidar        &  118.5 & 88.0  &{142.1} & 122.6 &  \textbf{208.3}\\
Assault       &  413.0 & 527.2 & {600.6} & 594.8 & \textbf{627.7}\\
BankHeist     &  97.7  & 34.2 & 131.6 & 266.7 & \textbf{508.0}\\
BattleZone    &  7833.3 & 5184.4 & 14870.0 & 15700.0 & \textbf{19400.0}\\
Breakout       &  2.3  & \textbf{16.4} & 4.9 & 1.8 & {3.2}\\
Freeway       &  28.7 & 20.3 & 26.7& 30.2 & \textbf{31.3}\\
Frostbite     &  1478.3 & 254.7 & 1181.3 & \textbf{2026.7} & 2010.0\\
Krull         &  3282.7 & {4539.9} &4229.6 &3171.9 & \textbf{3203.1}\\
MsPacman      &  1118.7 & 1480 & 1465.5 & 1482.3 & \textbf{1838.0}\\
Pong          &  -16.9 & \textbf{12.8} & -16.5 & -13.8 & {-14.2}\\
PrivateEye    &  97.8 & 58.3 & \textbf{218}.4 & 100 & {100}\\
Qbert         &  646.7 & 1288.8 & 1042.4 & {1830.8} & \textbf{2125}\\
RoadRunner    &  9923.3 & 5640.6 &5661.0 & {11913.3} & \textbf{15290.0}\\
Seaquest      &  396.0 & \textbf{683.3} & 384.5 & {570.7} & {458.0}\\
\bottomrule
\end{tabular}
\label{SUNRISE}
\end{table*}

\section{Experiments}
Through our experiments, we answer the following questions:
\begin{itemize}
    \item RQ1 : How does SEERL compare against traditional ensembles and SOTA reinforcement learning algorithms in terms of sample complexity and performance? 
    \item RQ2 : How does the diversity among policies contribute to the final performance?
    \item RQ3 : How does the policy selection framework help to find the best subset of policies?
    \item RQ4 : How are the policies obtained from SEERL any different from those obtained through random perturbation?
\end{itemize}
In the following sections, we describe the setup used for experimentation, the training procedure, the evaluation procedure and analysis of the results obtained. We answer the questions posed above in detail in the analysis section. 
\subsection{Setup}
In order to answer the above questions, we consider the environments from the Atari 2600 game suite \cite{bellemare2013arcade} for its discrete action space and Mujoco \cite{todorov2012mujoco} for its continuous action space. We conduct three sets of experiments to validate our framework :
\begin{itemize}
    \item The first set uses the following algorithms as the underlying reinforcement learning method (base learner) for SEERL - A2C \cite{DBLP:journals/corr/MnihBMGLHSK16}, ACER \cite{DBLP:journals/corr/WangBHMMKF16}, ACTKR \cite{DBLP:journals/corr/abs-1708-05144}, DDPG \cite{lillicrap2017continuous}, SAC \cite{haarnoja2018soft} and TRPO \cite{schulman2015trust}. The goal is to show that SEERL can be adapted to any reinforcement learning algorithm, both off-policy and on-policy. 
    \item The second set compares SEERL with three baseline ensemble methods that we have created. The three baselines are as follows:
    \begin{itemize}
    \item B1 : Ensembles of policies trained independently from a single algorithm. E.g., Five models of A2C.
    \item B2 : Ensembles of policies trained independently from different algorithms. E.g., two models of A2C, two models of ACER and one model from ACKTR.
    \item B3 : Ensembles of policies generated from random perturbation of model parameters at regular intervals. E.g., Five models of A2C, each of which has been obtained by perturbing at regular intervals and saving the parameters
    \end{itemize} 
    The goal is to understand how the traditional ensembles (B1 and B2) compare to SEERL and the role of diverse policies in an ensemble. Comparison with B3 is intended to show the importance of directed perturbation against random perturbation. 

    \item The third and final set compares SEERL with SOTA reinforcement learning algorithms that aim to be to sample efficient. We compare SEERL with SimPLE \cite{kaiser2019model}, CURL \cite{srinivas2020curl}, Rainbow DQN \cite{hessel2018rainbow} and SUNRISE \cite{lee2020sunrise}. In these experiments SEERL uses Rainbow DQN as the base learner in order to make a fair comparison with SUNRISE. The goal is to understand whether SEERL can be competitive with these algorithms that are sample efficient and be trained using the limited interactions it has with the environment. Comparison with SUNRISE is intended to understand whether SEERL is competitive 
    with a method that trains multiple critics in parallel. 
\end{itemize}
\subsection{Training}
SEERL uses a base learner to learn an ensemble of policies. This base learner can be any reinforcement learning algorithm such as A2C, SAC or Rainbow DQN. After selecting the base learner, SEERL is trained using the same hyper-parameters configurations as in the original implementation of the base learner.  The ambiguity that SEERL will lead to poor convergence as a result of shifting from zero to the maximum learning rate multiple times is mitigated through our results. SEERL performance during training is at least at par or better than the base learner, as shown in Figure 2(a, b). We train the models across different values of $M$ ranging from 3 to 9.\\
In order to make the comparison of SEERL with the base learner fair, we train both models on the same number of timesteps as specified in the original paper of the base learner. However, for comparison with SUNRISE, SimPLe, CURL and Rainbow DQN, we train SEERL using Rainbow DQN as base learner for 100k timesteps for a fair comparison.  Additional training results on the environments are presented in the supplementary under section 1. 
\subsection{Evaluation}
During evaluation, the policy selection framework is used to select the best subset of policies for the ensemble with $m=5$. All $m$ policies are loaded in parallel and provided with an observation from the environment. Based on the observation, every policy outputs an action, and the ensemble strategy decides the final action to be used on the environment. We perform this evaluation process for 100 episodes, and the average reward over these 100 episodes is reported in Table \ref{tab1} and Table \ref{tab2}. We use the SOTA scores from Rainbow DQN \cite{hessel2018rainbow} for benchmarking Atari 2600, and SEERL is seen to outperform it in many games, as seen in Table 1. For Mujoco environments, SEERL is seen to outperfom baselines by a considerable margin as shown in Table 2. Comparison of SEERL with baselines, B1 and B2, for Breakout and Half cheetah is shown in Figure 2(c,d). For comparison with SUNRISE, SimPLe, CURL and Rainbow DQN, the evaluation process is for 10 episodes and reported in Table \ref{SUNRISE}.

\begin{figure*}[t]
        \centering
        \begin{subfigure}{0.31\textwidth}
            \includegraphics[width=\textwidth]{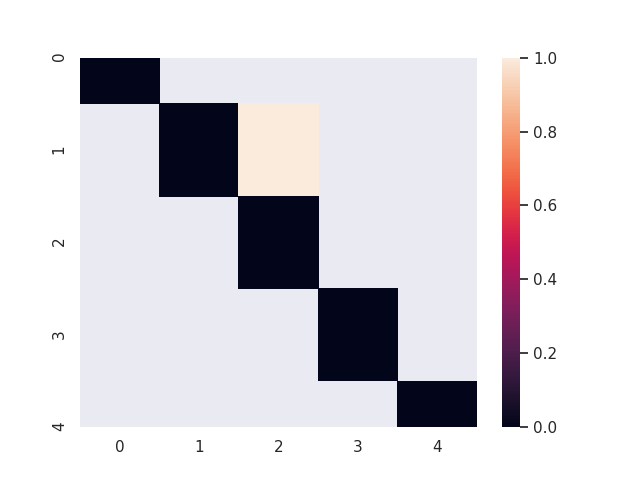}
            \caption{Divergence between independently trained policies used in the baseline, B1}
            \label{fig:gull6}
        \end{subfigure}
        \hfill
        \begin{subfigure}{0.31\textwidth}
            \includegraphics[width=\textwidth]{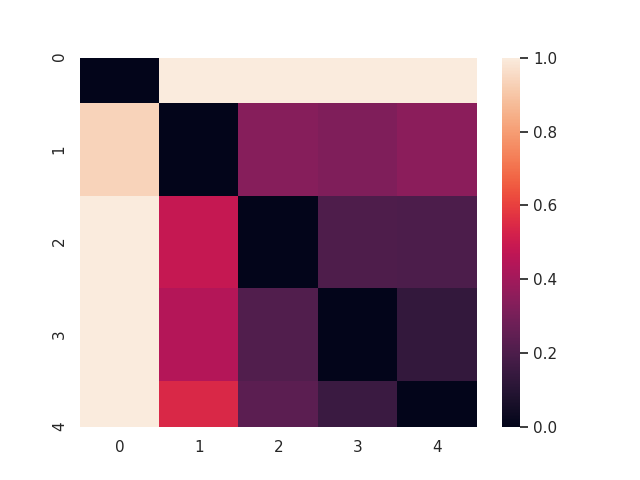}
            \caption{Divergence between SEERL policies}
            \label{fig:gull7}
        \end{subfigure}
        \hfill
        \begin{subfigure}{0.31\textwidth}
            \includegraphics[width=\textwidth]{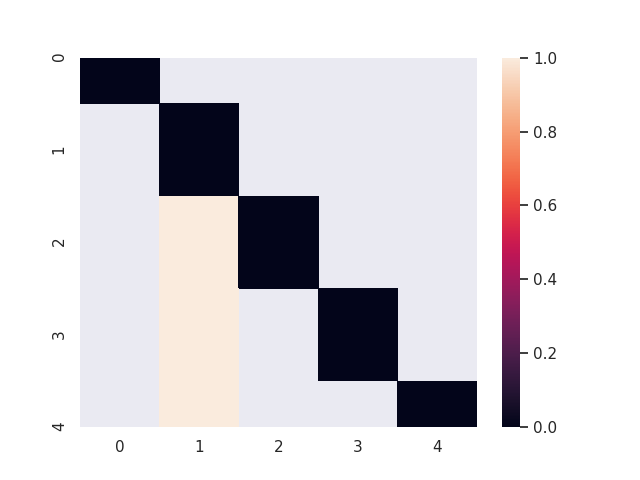}
            \caption{Divergence between policies obtained using random perturbations}
            \label{fig:gull8}
        \end{subfigure}
\end{figure*}


\begin{figure*}[t]
        \centering
        \begin{subfigure}{0.31\textwidth}
            \includegraphics[width=\textwidth]{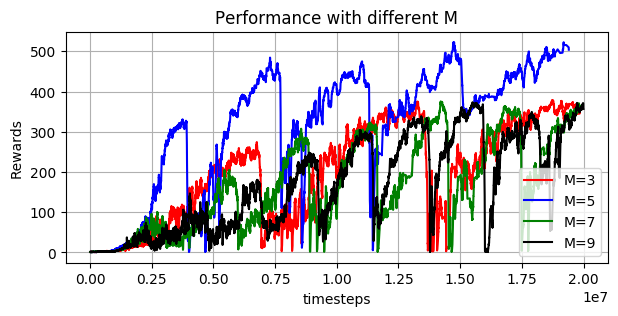}
            \caption{Training performance of SEERL as $M$ varies between 3 to 9}
            \label{fig:gull9}
        \end{subfigure}
        \hfill
        \begin{subfigure}{0.31\textwidth}
            \includegraphics[width=\textwidth]{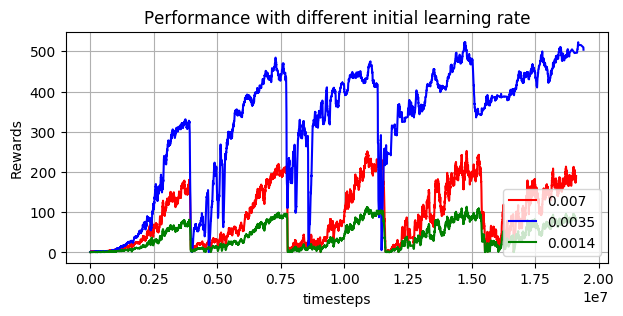}
            \caption{Training performance of SEERL as maximum learning rate $\alpha_0$ varies}
            \label{fig:gull10}
        \end{subfigure}
        \hfill
        \begin{subfigure}{0.31\textwidth}
            \includegraphics[width=\textwidth]{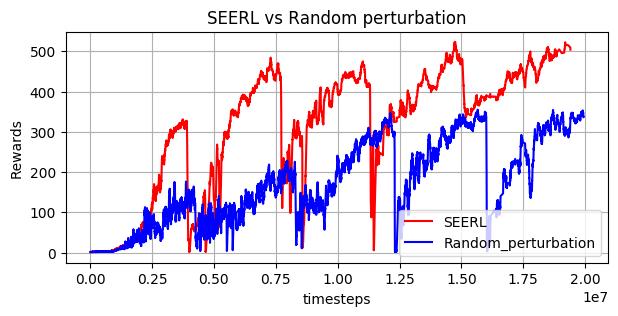}
            \caption{Training performance between SEERL and randomly perturbed model, with multiple perturbations in a sequence}
            \label{fig:gull11}
        \end{subfigure}
\end{figure*}

\subsection{Analysis}
We try to understand why and how SEERL gives such superior performance in comparison to baselines. We analyze the performance and sample efficiency of SEERL, the diversity among the policies, and finally, the comparison between a randomly perturbed model and SEERL. Analysis on individual performance of the SEERL policies and the dominance among policies in the ensemble are presented in the supplementary material under section 2.
\subsubsection{\textbf{RQ1 : Performance and sample efficiency of SEERL}}
The training curve in Figure 2(a,b) illustrates the performance of SEERL with the base learner and is observed to consistently outperform it. In Figure 2(a), it can be observed that SEERL achieves the same performance in 6 million timesteps that which requires the base learner to achieve in 20 million timesteps. Figure 2(c) and 2(d) illustrates the performance of SEERL in comparison with B1, B2 and is seen to outperform both baselines. Table \ref{tab1} shows that SEERL outperforms the  SOTA score from Rainbow DQN on most games. Similarly Table \ref{tab2} shows that SEERL outperforms across 5 of the 6 environments. The results in Table \ref{SUNRISE} illustrate comparison of SEERL with Rainbow DQN, SimPLe, CURL and SUNRISE at 100k timesteps. We observe that SEERL outperforms SUNRISE in 11 of the 15 Atari games used for comparison and achieves SOTA score for 100k interactions in 9 games. The performance of SEERL trained on 100k interactions highlights its sample efficiency. It is also observed that SEERL is 4-6 times faster than SUNRISE since it only uses a single network to sequentially train the policies while SUNRISE has multiple networks in parallel. As the number of networks grows in SUNRISE, the training time is longer while for SEERL, it remains the same. To summarize, SEERL outperforms most methods whether it is trained on millions of interactions or just 100k. 
\subsubsection{\textbf{RQ2 and RQ3 : Diversity of Policies and Policy Selection in SEERL}}
We establish the diversity among the individual policies concretely by computing the KL divergence between them using the action distribution across a diverse number of states. The higher the KL divergence between the policies, the more diverse the policies are. From Figure 3(b), we can observe that the SEERL policies are diverse, and diversity continues to exist as new models are formed. Conversely, for the baseline models, B1 and B3 (Figure 3(a) and Figure 3(c)), the KL divergence between the policies is substantial. This observation can be used to explain why the baseline ensembles failed to perform. The larger KL divergence among the baseline policies indicate that the policies do not have much overlap in the action space, and hence ensemble techniques such as majority voting were unable to find a good action. 
\\
Additionally, the analysis of the different subsets of policies in Figure 5 and their performance in Figure 2(e) confirms that diversity among policies is required, but too much diversity does not help. This empirically proves that extremely diverse models are not that helpful in an ensemble.
\\
We can, therefore conclude that SEERL can generate policies with sufficient diversity for a good ensemble.

\subsubsection{\textbf{RQ4: Random perturbation vs. SEERL}}
To emphasize that directed perturbation of the weights will lead to better models, we show the comparison between SEERL and a randomly perturbed model in Figure 4(c). This model has been perturbed with random values at regular intervals similarly to SEERL. The perturbation is done by backpropagating  random values of gradients instead of the actual values. We observe that doing a single perturbation or, multiple ones sequentially for a small period, does not improve performance. We can, therefore, establish that directed perturbation along the gradient direction is necessary to obtain a better model. In this experiment, all the hyper-parameters have been kept identical to that used in SEERL.


\begin{figure}[t]
        \centering
                \includegraphics[width=0.19\linewidth]{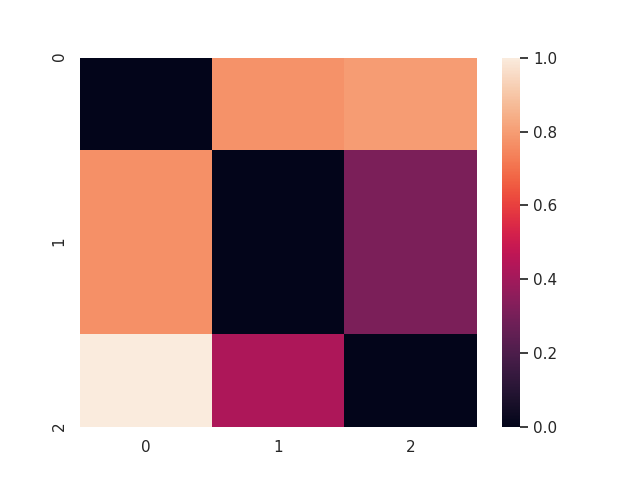}
                \includegraphics[width=0.19\linewidth]{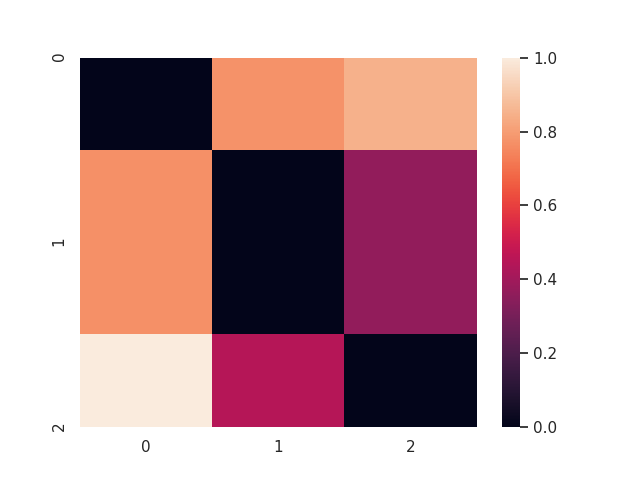}
                \includegraphics[width=0.19\linewidth]{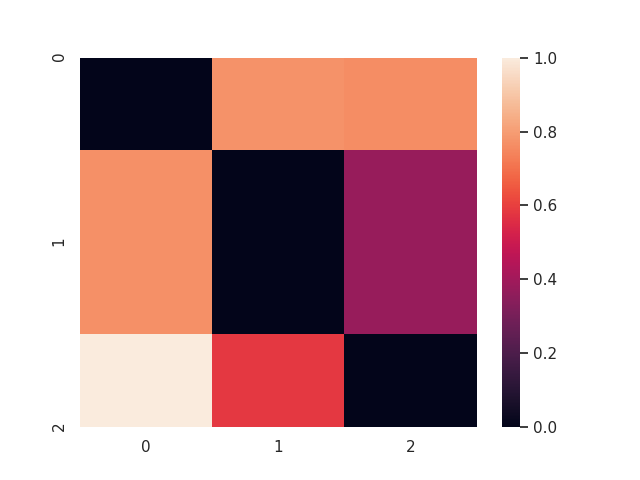}
                \includegraphics[width=0.19\linewidth]{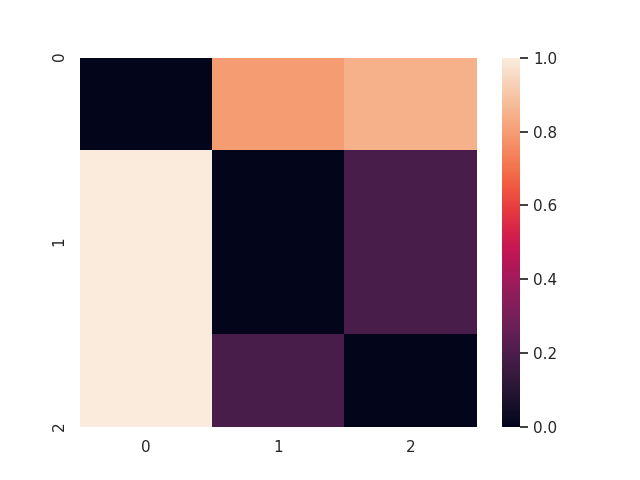}
                \includegraphics[width=0.19\linewidth]{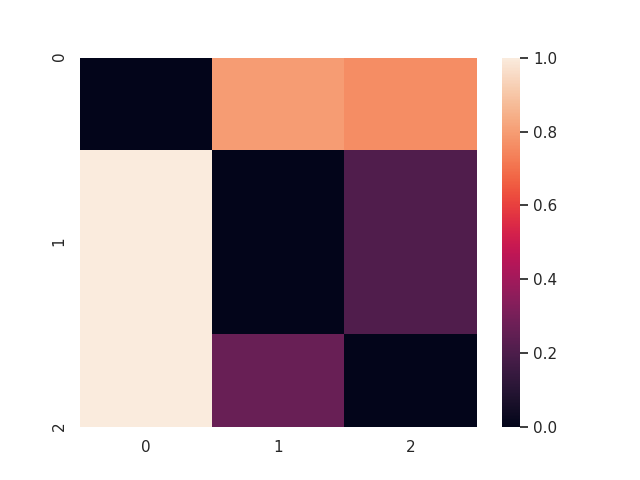}
        \\
                \includegraphics[width=0.19\linewidth]{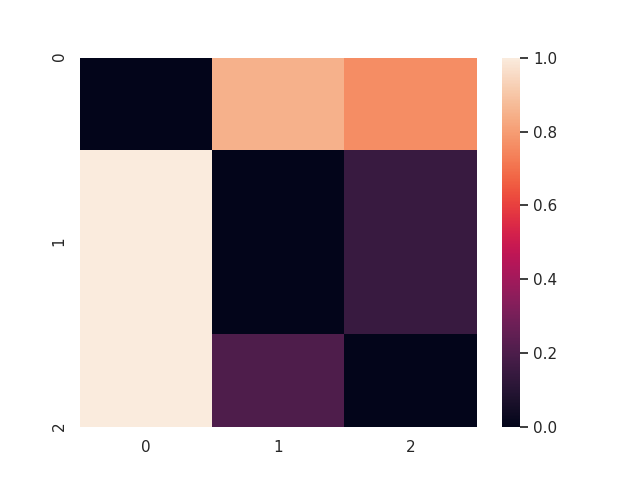}
                \includegraphics[width=0.19\linewidth]{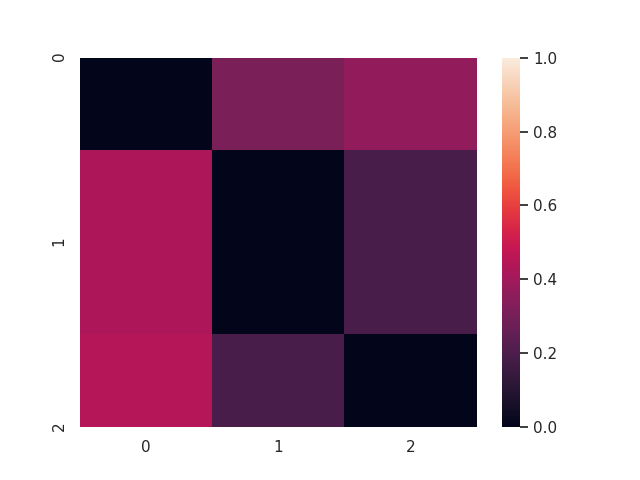}
                \includegraphics[width=0.19\linewidth]{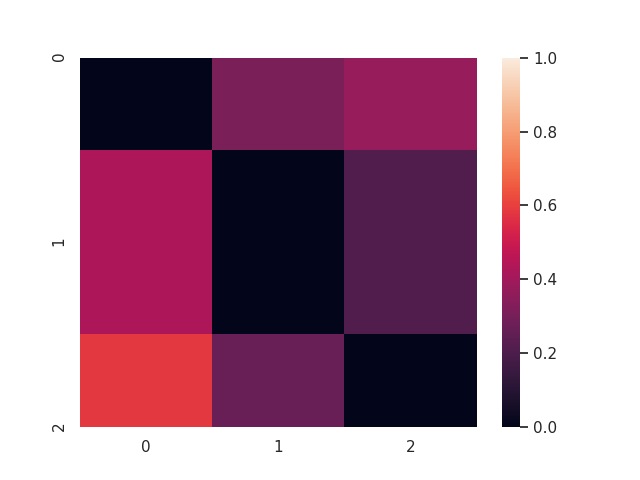}
                \includegraphics[width=0.19\linewidth]{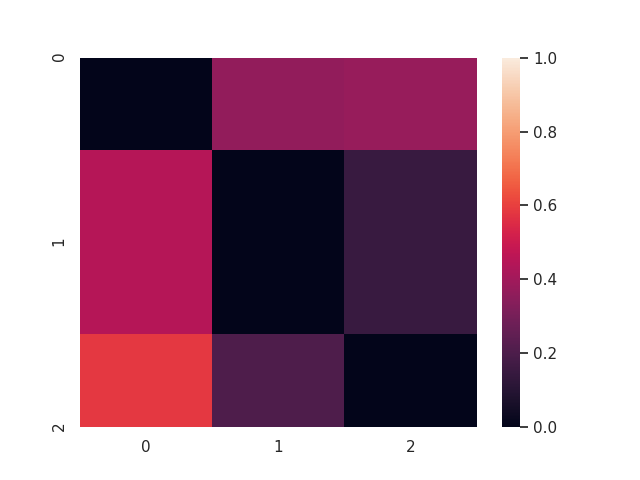}
                \includegraphics[width=0.19\linewidth]{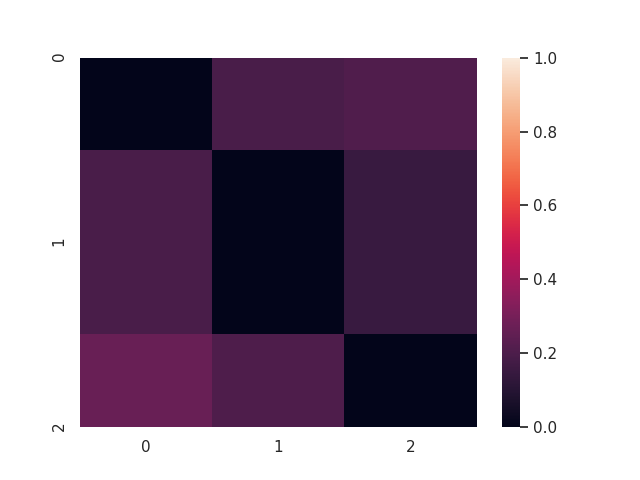}

    \caption{ Divergence between policies in a subset. Each subset consists of 3 policies chosen from a larger set of 5. The policies are trained on Breakout using A2C }
\end{figure}
\subsection{Ablation studies}
\subsubsection{\textbf{Effect of varying the number of cycles}}
The performance of SEERL is affected by the selection of $M$. For a fixed training budget, if the value of $M$ chosen is very large, the performance is seen to degrade. With larger $M$, the training cycle for each policy is reduced, thereby reducing the chance for the policy to settle at a good local minimum before it is perturbed again. In practice, we find that setting the value of $M$ between $3$ to $7$  works reasonably well. Figure 4(a) compares the performance of SEERL with varying $M$ values between $3$ and $9$. 
\subsubsection{\textbf{Effect of varying maximum learning rate}}
The maximum leaning rate value influences the performance of the policies and therefore affects the performance of SEERL. It directly impacts the perturbation of the local minima and hence, the diversity of the policies. In practice, we have seen that having a larger value tends to perform better, owing to the strong perturbation it causes at different local minima leading to reasonably different policies. We have used values ranging between $0.01$ to $0.001$ throughout our experiments. Figure 4(b)  compares the performance of SEERL with different values of $\alpha_0$ with $M=5$
\section{Conclusion and Future work}
In this paper, we introduce SEERL, a framework to ensemble multiple policies obtained from a single training run. We show that the policies learned at the different local minima are diverse in their performance and our policy selection framework helps to select the best subset of policies for the ensemble during evaluation. SEERL outperforms the three baseline methods and beats SOTA scores in complex environments having discrete and continuous action spaces. We show our results using both off-policy and on-policy reinforcement learning algorithms and therefore showcase the scalability of the framework. Our analysis shows that, in comparison to baselines, SEERL achieves comparable, and sometimes better performance using a significantly low number of samples, making it an extremely sample efficient algorithm. 
Future work will explore how to combine the learned policies during training time as a growing ensemble to stabilize training and increase diversity.


\newpage
\bibliographystyle{ACM-Reference-Format} 
\balance
\bibliography{sample}


\end{document}